\documentclass[a4paper,twoside]{article}
\usepackage{booktabs}
\usepackage{epsfig}
\usepackage{subcaption}
\usepackage{calc}
\usepackage{amssymb}
\usepackage{amstext}
\usepackage{amsmath,bm}

\usepackage{amsthm}
\usepackage{multicol}
\usepackage{pslatex}
\usepackage[hidelinks]{hyperref}
\usepackage{verbatim}
\usepackage{multirow}
\usepackage[bottom]{footmisc}
\usepackage[margin=0.5in]{geometry}
\usepackage{SCITEPRESS}     
\usepackage{natbib}
\usepackage{balance}
\usepackage{acronym}
\usepackage{lscape}
\usepackage{graphicx} 
\usepackage{caption} 
\usepackage{tabularx}
\usepackage{algorithm}
\usepackage{algorithmic}
\usepackage[normalem]{ulem}
\useunder{\uline}{\ul}{}
\usepackage{tikz}
\usepackage{adjustbox}
\usepackage{fontawesome}

\usepackage{stfloats}
\usetikzlibrary{positioning, arrows.meta, calc}

\begin{document}
\acrodef{AI}{Artificial Intelligence}

\acrodef{FL}{Federated Learning}

\acrodef{IID}{Independent and Identically Distributed}

\acrodef{PSO}{Particle Swarm Optimization}

\acrodef{CNN}{Convolutional Neural Network}

\acrodef{CNNs}{Convolutional Neural Networks}

\acrodef{CT}{Computed Tomography}

\acrodef{non-IID}{non-independent and identically distributed}

\acrodef{Biglycan}{Breast Cancer Histopathological Dataset}

\acrodef{GANs}{Generative Adversarial Networks}

\acrodef{LC25000}{Lung and Colon Cancer Histopathological Images}

\acrodef{IHC}{Immunohistochemistry}

\acrodef{HPO}{hyperparameter optimization}

\acrodef{LR}{Learning Rate}

\acrodef{MRI}{Magnetic Resonance Imaging}

\acrodef{ResNet}{Residual Network}

\acrodef{AUC}{Area Under the Curve}

\acrodef{FedAvg}{Federated Averaging}

\acrodef{TPE}{Tree-structured Parzen Estimator}

\title{Generalizable Hyperparameter Optimization for Federated Learning on Non-IID Cancer Images}


\author{\authorname{Elisa Gonçalves Ribeiro\sup{1}\orcidAuthor{0009-0001-0882-1685}, Rodrigo Moreira\sup{1}\orcidAuthor{0000-0002-9328-8618}, \\Larissa Ferreira {Rodrigues Moreira}\sup{1}\orcidAuthor{0000-0001-8947-9182}, André Ricardo Backes\sup{2}\orcidAuthor{0000-0002-7486-4253}}
\affiliation{\sup{1}Institute of Exact and Technological Sciences, Federal University of Viçosa - UFV, Rio Paranaíba-MG, Brazil}
\affiliation{\sup{2}Department of Computing, Federal University of São Carlos, São Carlos-SP, Brazil}
\email{\{elisa.ribeiro, rodrigo, larissa.f.rodrigues\}@ufv.br, arbackes@yahoo.com.br}
}


\abstract{
Deep learning for cancer histopathology training conflicts with privacy constraints in clinical settings. Federated Learning (FL) mitigates this by keeping data local; however, its performance depends on hyperparameter choices under non-independent and identically distributed (non-IID) client datasets. This paper examined whether hyperparameters optimized on one cancer imaging dataset generalized across non-IID federated scenarios. We considered binary histopathology tasks for ovarian and colorectal cancers. We perform centralized Bayesian hyperparameter optimization and transfer dataset-specific optima to the non-IID FL setup. The main contribution of this study is the introduction of a simple cross-dataset aggregation heuristic by combining configurations by averaging the learning rates and considering the modal optimizers and batch sizes. This combined configuration achieves a competitive classification performance.
}

\onecolumn \maketitle \normalsize \setcounter{footnote}{0} \vfill

\acrodef{AI}{Artificial Intelligence}

\acrodef{FL}{Federated Learning}

\acrodef{IID}{Independent and Identically Distributed}

\acrodef{PSO}{Particle Swarm Optimization}

\acrodef{CNN}{Convolutional Neural Network}

\acrodef{CNNs}{Convolutional Neural Networks}

\acrodef{CT}{Computed Tomography}

\acrodef{non-IID}{non-independent and identically distributed}

\acrodef{Biglycan}{Breast Cancer Histopathological Dataset}

\acrodef{GANs}{Generative Adversarial Networks}

\acrodef{LC25000}{Lung and Colon Cancer Histopathological Images}

\acrodef{IHC}{Immunohistochemistry}

\acrodef{HPO}{hyperparameter optimization}

\acrodef{LR}{Learning Rate}

\acrodef{MRI}{Magnetic Resonance Imaging}

\acrodef{ResNet}{Residual Network}

\acrodef{AUC}{Area Under the Curve}

\acrodef{FedAvg}{Federated Averaging}

\acrodef{TPE}{Tree-structured Parzen Estimator}

\section{\uppercase{Introduction}}
Cancer remains a leading global health burden, with projections of approximately 28.4 million new cases by 2040, and 35 million by 2050, representing increases of approximately 47\% and 77\%, respectively. Colorectal and ovarian cancers are particularly relevant, with colorectal cancer accounting for more than 1.9 million new cases and 930,000 deaths worldwide annually. Ovarian cancer causes approximately 314,000 new cases and 207,000 deaths annually and is frequently diagnosed at advanced stages, leading to high mortality rates~\citep{Morgan2022, Roshandel2024, Bray2024}.  

Deep learning has arisen as an approach for automated cancer diagnosis based on medical images. In histopathology, \ac{CNNs} and other deep architectures, often initialized with ImageNet weights, can achieve expert-level performance in distinguishing cancerous from non-cancerous tissues~\citep{RodriguesMoreira2025jdi}. However, the standard centralized training paradigm, in which large image collections are pooled in a single repository, conflicts with privacy regulations and institutional data governance, and the fact that data are naturally distributed across hospitals and laboratories~\citep{Guan2024}. 

\ac{FL} has emerged as an alternative that enables collaborative training without sharing raw data. Institutions keep data on-site and periodically send model updates to a coordinating server, which aggregates them into a global server~\citep{McMahan2017, Leonardo2025}. \ac{FL} has been applied to several medical imaging tasks, including brain tumor segmentation, chest X-ray classification, and computational pathology, achieving a performance close to centralized training while preserving data locality~\citep{Guan2024, Moreira2025, Barbosa2025}. 

However, practical deployment in clinical environments remains challenging because institutional datasets are rarely independent and identically distributed (IID). They differ in size, class balance, acquisition protocols, and patient demographics, which leads to strong \ac{non-IID} client partitions and can degrade convergence and generalization~\citep{Leonardo2025}. In addition, \ac{FL} models are sensitive to hyperparameter tuning, and hyperparameters tuned for one site or dataset may transfer poorly to others, particularly under \ac{non-IID} conditions. Existing studies on \ac{HPO} for FL often focus on a single dataset or assume fixed, task-specific configurations~\citep{Kuo2023, Kundroo2024}; thus, obtaining hyperparameter configurations that generalize across \ac{non-IID} federated cancer-imaging scenarios remains an open question.

In this paper, we address this gap by studying whether hyperparameters optimized on one cancer imaging dataset can be generalized across \ac{non-IID} federated scenarios and across tumor types. We designed a two-phase pipeline that (i) performs centralized Bayesian \ac{HPO} with the \ac{TPE} on two binary histopathology tasks (ovarian and colon cancer) and (ii) transfers and combines the resulting configurations in a \ac{non-IID} FL setup based on \ac{FedAvg}. By comparing multiple CNN architectures under ovary-optimized, colon-optimized, and combined hyperparameter schemes, we assessed how these choices affect federated performance and whether simple combined configurations can act as robust strategies that generalize across \ac{non-IID} federated cancer scenarios.


\section{\uppercase{Related Work}}\label{sec:rw}

In this section we review prior work at the intersection of cancer imaging, \ac{HPO} and federated or privacy preserving learning. 
\cite{Subramanian2022} addresses privacy-preserving classification of collaborative cancer models across heterogeneous, \ac{non-IID} datasets. The authors propose a decentralized \ac{FL} framework using FedAvg and FedProx, coupled with client-server Bayesian \ac{HPO} and a selective update rule based on local performance improvement. Evaluating MobileNetV3, XceptionV3, and ResNet201 on CT and MRI data of cervical, lung, and colon cancers, they report accuracy, precision, recall, and F1 scores across varying communication rounds.

\cite{Khan2024} addresses automated breast cancer detection under privacy constraints and data scarcity. They proposed a decentralized federated framework where hospitals train \ac{CNNs} using an ant colony optimization-based bilevel strategy for joint hyperparameter and architecture search, exchanging updates via a circular topology. Using the DDSM dataset distributed across three clients for binary classification, they show that the ant colony scheme outperforms particle swarm optimization (PSO) and grid search baselines in terms of accuracy.

\cite{Tamanini2025} focuses on efficient endometrioid ovarian carcinoma detection in histopathology images with limited data. The pipeline extracts radiomics features from hematoxylin and eosin images converted into eight color spaces, applying incremental recursive feature elimination before training traditional classifiers and CNN baselines. The study reports that the YCrCb color space with Random Forest achieves nearly 98\% accuracy.

\cite{Yang2025} addresses data-scarce skin cancer detection in IoT healthcare settings. The authors propose an active learning framework combining a CNN classifier with a reinforcement learning sample selector and an enhanced artificial bee colony algorithm for \ac{HPO}. Evaluated on ISIC, HAM10000, and PH2 datasets, the approach shows consistent gains in accuracy and memory usage compared to various machine learning and deep learning baselines.

\cite{Li2025} optimizes hyperparameters for breast cancer classification to overcome the instability of existing metaheuristics. The authors introduce the Multi-Strategy Parrot Optimizer (MSPO), enhanced with Sobol initialization and chaotic perturbations, to tune a ResNet18 model. Benchmarked against CEC 2022 functions and applied to the BreakHis dataset, the MSPO-tuned model outperforms untuned versions and alternative algorithms in accuracy, precision, recall, and F1 score.

\cite{Almelibari2025} tackles diabetes prediction while addressing privacy, class imbalance, and communication costs. The framework combines a tree-based and neural ensemble with a PSO-based weighting scheme, extended into a privacy-preserving federated architecture. Using SMOTE for class balancing on a public dataset, the study reports that their proposed weighted aggregation yields higher accuracy and communication efficiency. 

\begin{table*}[htbp]
\caption{Short comparison of the Related Works.}
\label{tab:related-work}
\resizebox{\textwidth}{!}{%
\renewcommand{\arraystretch}{1.2}
\begin{tabular}{@{}lcccccccc@{}}
\toprule
\textbf{Approach}      & \textbf{\begin{tabular}[c]{@{}c@{}}Cancer \\ Task\end{tabular}} & \textbf{\begin{tabular}[c]{@{}c@{}}Medical \\ Imaging\end{tabular}} & \textbf{\begin{tabular}[c]{@{}c@{}}Histopathology \\ Images\end{tabular}} & \textbf{\begin{tabular}[c]{@{}c@{}}Federated \\ Learning\end{tabular}} & \textbf{\begin{tabular}[c]{@{}c@{}}Non-IID \\ Clients\end{tabular}} & \textbf{\begin{tabular}[c]{@{}c@{}}HPO \\ Focus\end{tabular}} & \textbf{\begin{tabular}[c]{@{}c@{}}HPO in \\ FL\end{tabular}} & \textbf{\begin{tabular}[c]{@{}c@{}}Multiple \\ Cancer Types\end{tabular}} \\ \midrule
\cite{Subramanian2022} & \faCircle                                                       & \faCircle                                                           & \faCircleO                                                                & \faCircle                                                              & \faCircle                                                           & \faCircle                                                     & \faCircle                                                     & \faCircle                                                                 \\
\cite{Khan2024}        & \faCircle                                                       & \faCircle                                                           & \faCircleO                                                                & \faCircle                                                              & \faCircleO                                                          & \faCircle                                                     & \faCircle                                                     & \faCircleO                                                                \\
\cite{Tamanini2025}    & \faCircle                                                       & \faCircle                                                           & \faCircle                                                                 & \faCircleO                                                             & \faCircleO                                                          & \faCircleO                                                    & \faCircleO                                                    & \faCircleO                                                                \\
\cite{Yang2025}        & \faCircle                                                       & \faCircle                                                           & \faCircleO                                                                & \faCircleO                                                             & \faCircleO                                                          & \faCircle                                                     & \faCircleO                                                    & \faCircleO                                                                \\
\cite{Li2025}          & \faCircle                                                       & \faCircle                                                           & \faCircle                                                                 & \faCircleO                                                             & \faCircleO                                                          & \faCircle                                                     & \faCircleO                                                    & \faCircleO                                                                \\
\cite{Almelibari2025}          & \faCircleO                                                      & \faCircle                                                          & \faCircleO                                                                & \faCircle                                                              & \faCircleO                                                          & \faCircle                                                     & \faCircle                                                     & \faCircleO                                                                \\ \midrule
\textbf{Our Approach}  & \faCircle                                                       & \faCircle                                                           & \faCircle                                                                 & \faCircle                                                              & \faCircle                                                           & \faCircle                                                     & \faCircle                                                     & \faCircle                                                                 \\ \bottomrule
\end{tabular}%
}
\end{table*}

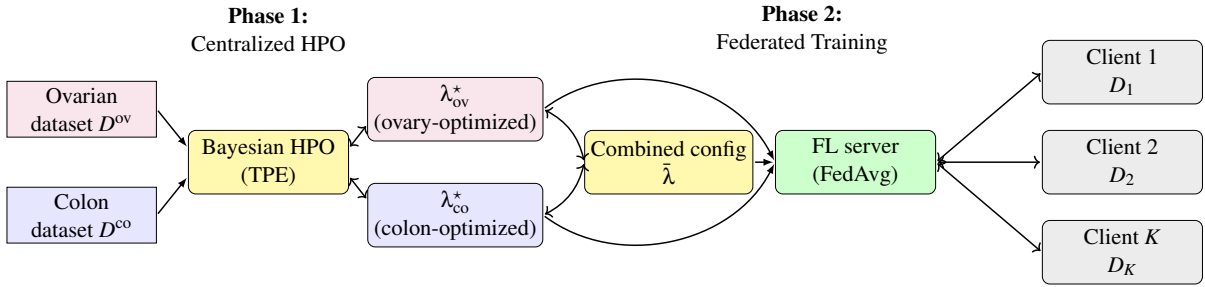
\begin{figure*}[!ht]
\resizebox{\textwidth}{!}{%
\centering
\begin{tikzpicture}[
    font=\large,
    node distance=2cm and 2.5cm, 
    block/.style={draw, rounded corners, align=center, minimum width=3cm, minimum height=1.2cm},
    ds/.style={draw, align=center, minimum width=2.8cm, minimum height=1cm},
    line/.style={-latex, thick, shorten <=1pt, shorten >=1pt}, 
    line_bi/.style={<->, thick, shorten <=1pt, shorten >=1pt} 
]

\node[align=center] (title1) at (-4.0,4.0) {\textbf{Phase 1:}\\Centralized HPO};
\node[align=center] (title2) at (6.0,4.0) {\textbf{Phase 2:}\\Federated Training};

\node[ds, fill=purple!10] (ov) at (-7.5,2.5) {Ovarian\\dataset $D^{\mathrm{ov}}$};
\node[ds, fill=blue!10] (co) at (-7.5,0.5) {Colon\\dataset $D^{\mathrm{co}}$};

\node[block, fill=yellow!40] (hpo) at (-4.0,1.5) {Bayesian HPO\\(TPE)};
\draw[line] (ov.east) -- (hpo.170);
\draw[line] (co.east) -- (hpo.190);

\node[block, fill=purple!10] (lambda_ov) at (-0.5,2.5) {$\lambda^{\star}_{\mathrm{ov}}$\\(ovary-optimized)};
\node[block, fill=blue!10] (lambda_co) at (-0.5,0.5) {$\lambda^{\star}_{\mathrm{co}}$\\(colon-optimized)};
\draw[line_bi] (hpo.10) -- (lambda_ov.190); 
\draw[line_bi] (hpo.-10) -- (lambda_co.170);

\node[block, fill=yellow!40] (comb) at (3.5,1.5) {Combined config\\$\bar{\lambda}$};
\draw[line_bi] (lambda_ov.east) to[bend left=25] (comb.west);
\draw[line_bi] (lambda_co.east) to[bend right=25] (comb.west);

\node[block, fill=green!20] (server) at (7,1.5) {FL server\\(FedAvg)};
\draw[line] (lambda_ov.east) to[bend left=50] (server.west);
\draw[line] (lambda_co.east) to[bend right=50] (server.west);
\draw[line] (comb.east) -- (server.west);

\node[block, fill=darkgray!10] (c2) at (12,1.5) {Client 2\\$D_2$};
\node[block, fill=darkgray!10] (c1) at (12,3.2) {Client 1\\$D_1$}; 
\node[block, fill=darkgray!10] (cK) at (12,-0.2) {Client $K$\\$D_K$};

\draw[line_bi] (server.east) -- (c1.west);
\draw[line_bi] (server.east) -- (c2.west);
\draw[line_bi] (server.east) -- (cK.west);

\end{tikzpicture}
}
\caption{Proposed Method. Phase~1 performs centralized Bayesian hyperparameter optimization with TPE on ovarian and colon datasets, yielding task-specific optima $\lambda^{\star}_{\mathrm{ov}}$ and $\lambda^{\star}_{\mathrm{co}}$ and a combined configuration $\bar{\lambda}$. Phase~2 evaluates these three configurations in a \ac{non-IID} federated setting using \ac{FedAvg}.}
\label{fig:method_overview}
\end{figure*}

We summarize our contribution in Table~\ref{tab:related-work}, 
where we use {\scriptsize \faCircle} when the approach addresses the feature 
and {\scriptsize \faCircleO} otherwise. 

\textbf{Contribution Positioning.} While all solutions in Table~\ref{tab:related-work} focus on improving cancer prediction or clinical decision support either by designing new \ac{FL} protocols, proposing metaheuristic hyperparameter optimizers, or building more accurate centralized imaging pipelines, none of them investigates whether hyperparameters tuned on one dataset can generalize across \ac{non-IID} federated cancer imaging scenarios and across tumor types. In this paper, we propose a two stage methodology that first performs Bayesian hyperparameter optimization on centralized histopathology tasks for colon and ovarian cancer and then transfers and combines the resulting configurations in a \ac{FL} setup, enabling a controlled comparison across different \ac{CNNs} and providing empirical evidence and practical guidelines on hyperparameter schemes that behave robustly under heterogeneous federated conditions.

\section{\uppercase{Material and Methods}}\label{sec:method}

Figure~\ref{fig:method_overview} summarizes the proposed method. In Phase~1, we perform centralized Bayesian \ac{HPO} with \ac{TPE} on two binary histopathology tasks (ovarian and colon), obtaining dataset specific optima $\lambda^{\star}_{\mathrm{ov}}$ and $\lambda^{\star}_{\mathrm{co}}$ and constructing a combined configuration $\bar{\lambda}$ from them. In Phase~2, we instantiate a \ac{non-IID} \ac{FL} scenario with $K$ clients trained with \ac{FedAvg}, and evaluate these three hyperparameter schemes under identical model architectures and data partitions. 

\subsection{Medical Imaging Datasets}

We consider two binary histopathology classification tasks, both framed as ``cancer'' vs. ``non cancer''. The first dataset is the Ovarian Cancer \& Subtypes Histopathology dataset\footnote{\url{https://www.kaggle.com/datasets/bitsnpieces/ovarian-cancer-and-subtypes-dataset-histopathology}}, which contains digital pathology patches extracted from hematoxylin and eosin (H\&E) stained ovarian tissue slides. We group the available subtypes into a single positive class (ovarian cancer) and use benign or non tumoral tissue as the negative class. 

\begin{figure}[!ht]
    \centering
    \resizebox{\columnwidth}{!}{
    \begin{tabular}{ccc}
        \includegraphics[width=0.5\columnwidth]{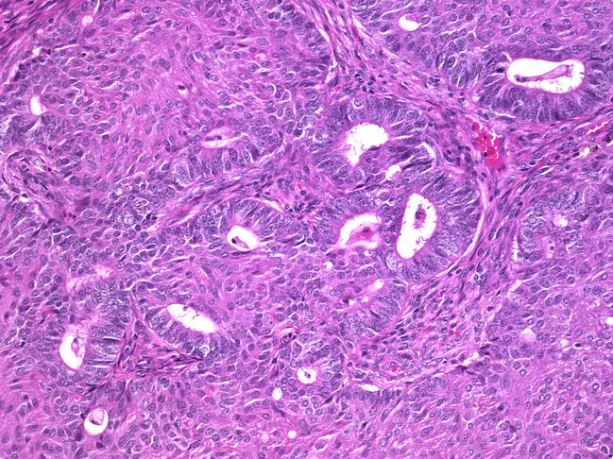} &
        \includegraphics[width=0.5\columnwidth]{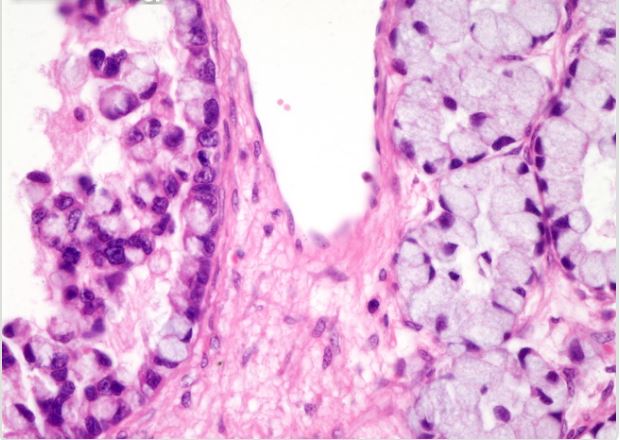} &
        \includegraphics[width=0.5\columnwidth]{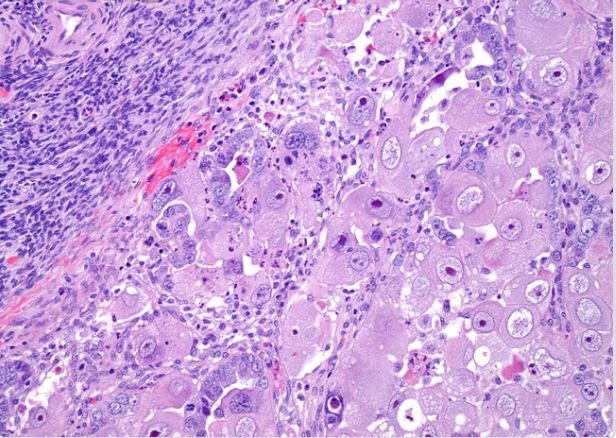} \\
        \multicolumn{3}{c}{\LARGE(a) Ovarian cancer} \\[4pt]
        
        \includegraphics[width=0.5\columnwidth]{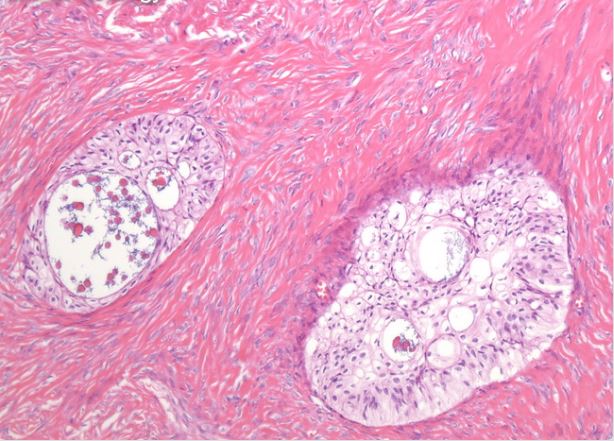} &
        \includegraphics[width=0.5\columnwidth]{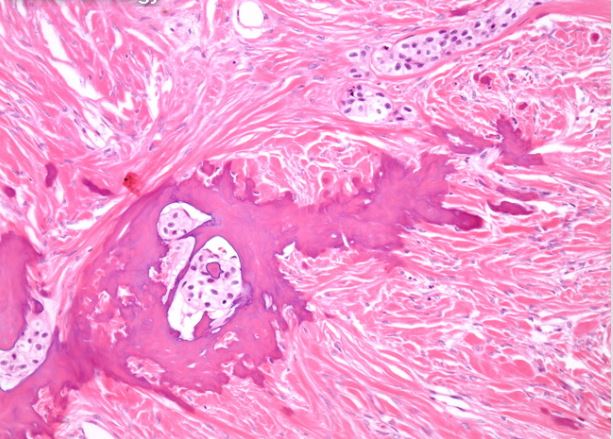} &
        \includegraphics[width=0.5\columnwidth]{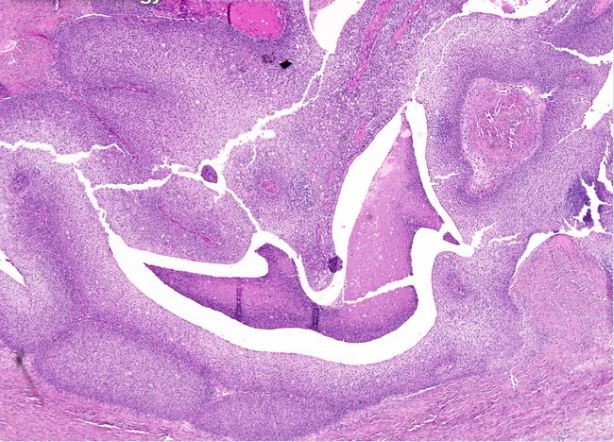} \\
        \multicolumn{3}{c}{\LARGE(b) Ovarian normal} \\[4pt]
        
        \includegraphics[width=0.5\columnwidth]{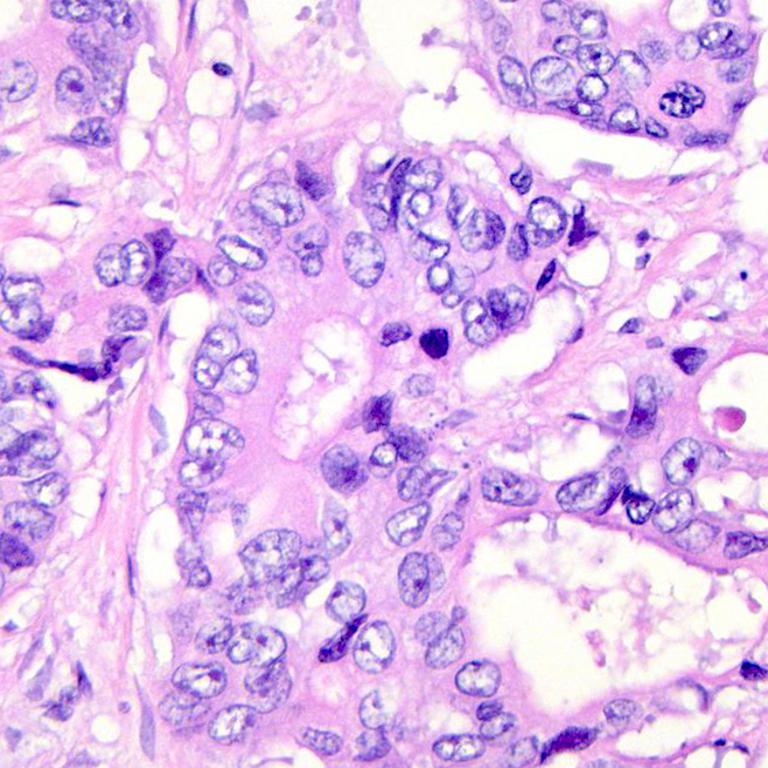} &
        \includegraphics[width=0.5\columnwidth]{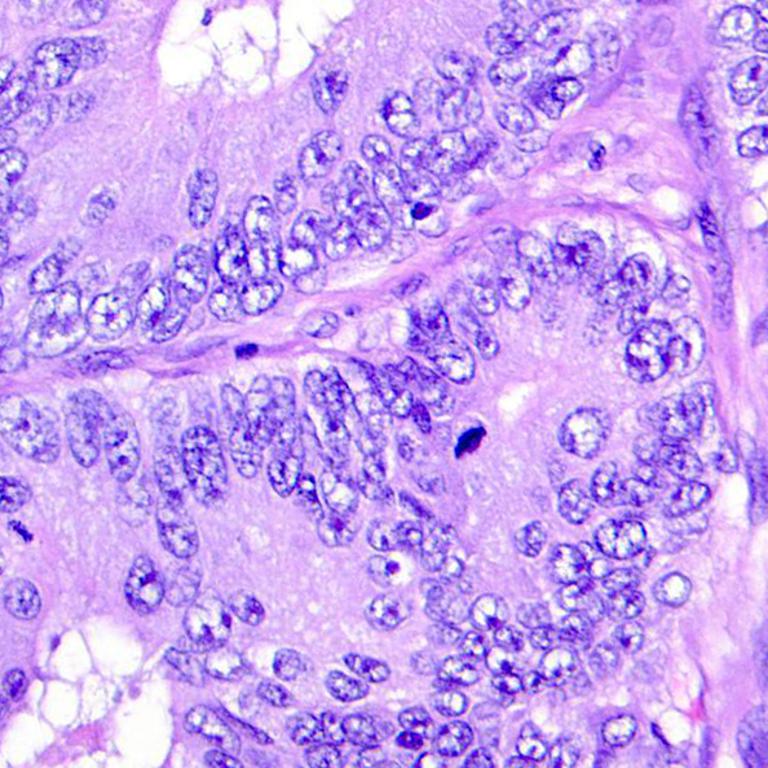} &
        \includegraphics[width=0.5\columnwidth]{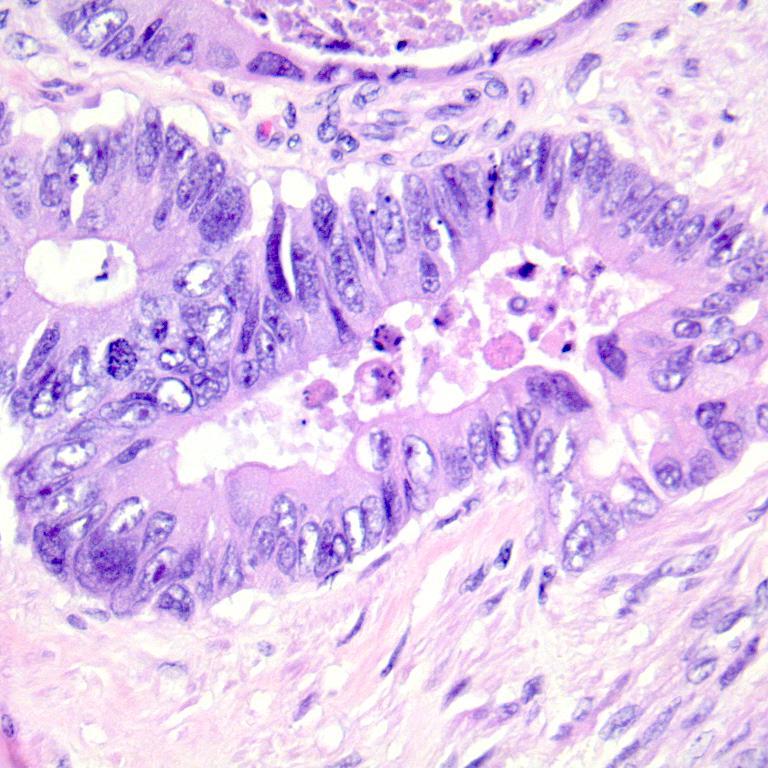} \\
        \multicolumn{3}{c}{\LARGE(c) Colon cancer} \\[4pt]
        
        \includegraphics[width=0.5\columnwidth]{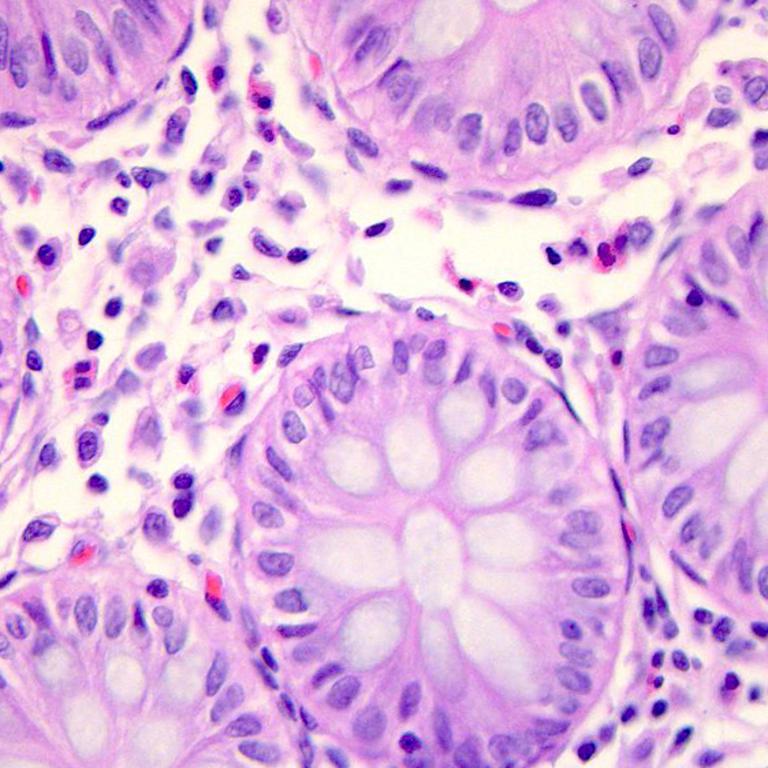} &
        \includegraphics[width=0.5\columnwidth]{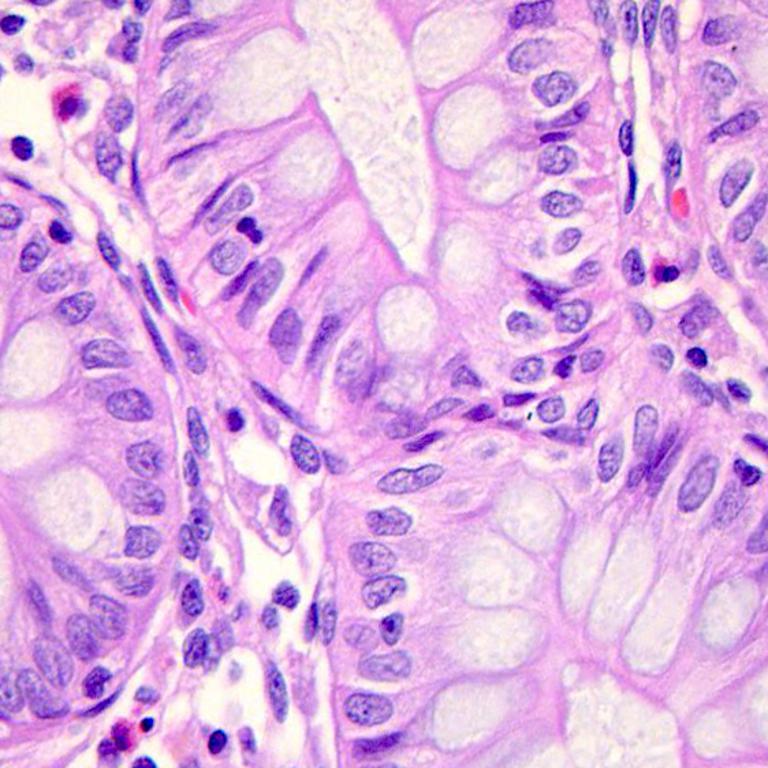} &
        \includegraphics[width=0.5\columnwidth]{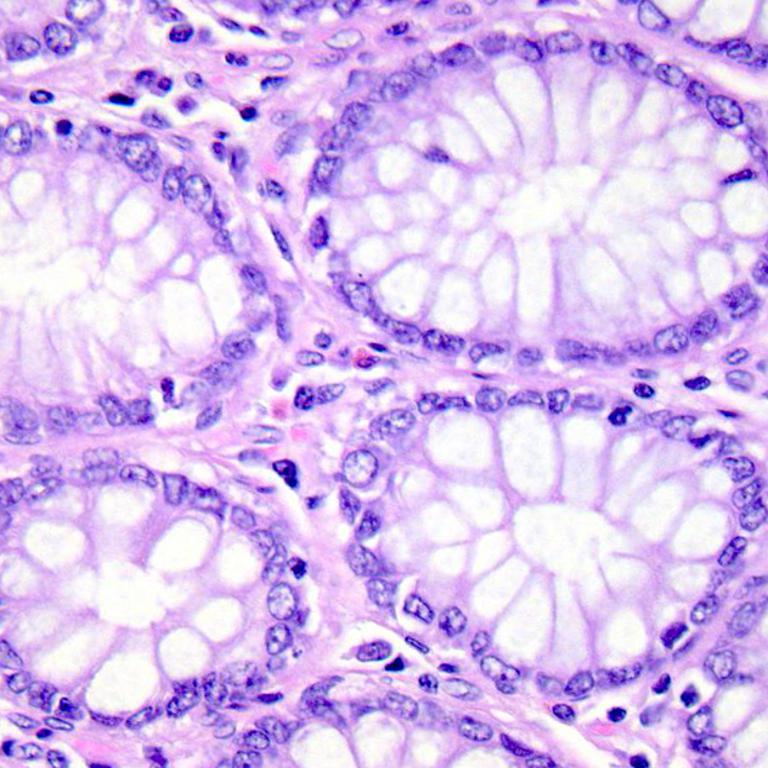} \\
        \multicolumn{3}{c}{\LARGE(d) Colon normal} \\             
    \end{tabular}
    }
    \caption{Examples of images from the two datasets used in this study.}
    \label{fig:dataset_examples}
\end{figure}

The second dataset is a subset of the LC25000 dataset\footnote{\url{https://www.kaggle.com/andrewmvd/lung-and-colon-cancer-histopathological-images}}, which contains colon histopathology images with normal and tumoral samples. We select only colon images and map them to a binary label (cancer vs. non cancer). Representative examples of ovarian and colon patches for both classes (Figure~\ref{fig:dataset_examples}).

To enable a controlled comparison, we sub sample the LC25000 colon subset so that the resulting colon dataset matches the ovarian dataset in terms of total number of images and class imbalance ratio. After this alignment, the combined dataset comprises 996 images (ovary + colon). For each dataset, we randomly split the images into 80\% for training and 20\% for testing, with a stratified split on the binary label. Within the training portion, we further reserve a validation subset that is used exclusively for \ac{HPO}. All images are resized to 224 $\times$ 224 pixels and normalized with ImageNet statistics. 

\subsection{CNN Architectures}
\label{subsec:cnns}

We evaluate six \ac{CNNs} that span from classic, low-depth models to more modern, parameter-efficient networks. All models are initialized with ImageNet pretrained weights and adapted to binary classification by replacing the final fully connected layer.

\begin{itemize}
    \item \textbf{AlexNet}: a CNN with five convolutional layers and three fully connected layers, included as a classical baseline for comparison with more recent architectures~\citep{Krizhevsky2012}.
    
    \item \textbf{ResNet-18 / ResNet-34 / ResNet-50}: residual networks with increasing depth that use skip connections to ease optimization. ResNet-18 and ResNet-34 represent moderately deep backbones, while ResNet-50 provides a higher-capacity model built from bottleneck blocks~\citep{He2016}.
    
    \item \textbf{SqueezeNet}: a compact architecture based on ``fire'' modules that reduces the number of parameters. It is included to assess whether lightweight models can achieve competitive performance in the federated histopathology setting~\citep{Leonardo2025}.
    
    \item \textbf{EfficientNet-B0}: a modern CNN obtained by compound scaling of depth, width, and resolution. It provides a strong parameter-efficient baseline and is particularly suited to capturing fine-grained texture patterns in histopathology images~\citep{Tan2019}.
\end{itemize}

\subsection{Federated Learning Pipeline}

We adopt a two phase pipeline that combines centralized \ac{HPO} and federated training. In the first phase, we perform centralized training and evaluation on each dataset separately. Let $D^{\mathrm{ov}}$ denote the ovarian dataset and $D^{\mathrm{co}}$ the colon dataset. For each dataset, we fine tune ImageNet pretrained \ac{CNN}s with a task specific classification head and use the validation split to drive \ac{HPO}, as detailed in Section~\ref{subsec:hpo_tpe}.

In the second phase, we simulate a cross silo federated learning scenario with $K$ logical clients. The training data from $D^{\mathrm{ov}}$ and $D^{\mathrm{co}}$ are partitioned into $K$ local datasets $\{D_k\}_{k=1}^{K}$ under a \ac{non-IID} scheme that reflects realistic heterogeneity in clinical environments. Each client $k$ receives a subset $D_k$ characterized by different sample sizes and label proportions, so that the empirical label distribution $p_k(y)$ varies across clients. Federated training proceeds in communication rounds indexed by $t = 1, \dots, T$. At round $t$, the server broadcasts the current global parameters $\theta^{(t)}$ to all (or a subset of) clients, each client performs local stochastic gradient descent (SGD) updates for a fixed number of epochs on its local data, and the resulting local parameters $\{\theta_k^{(t+1)}\}_{k=1}^{K}$ are aggregated on the server using \ac{FedAvg}~\citep{McMahan2017} (Equation: \ref{eq:fedavg}).
\begin{equation}\label{eq:fedavg}
    \theta^{(t+1)} = \sum_{k=1}^{K} \frac{n_k}{\sum_{j=1}^{K} n_j} \, \theta_k^{(t+1)},
\end{equation}
where $n_k = |D_k|$ is the number of training samples at client $k$. All federated experiments use the same \ac{CNN} architectures and differ only in the hyperparameter configuration chosen for learning rate, optimizer, and batch size.

\subsection{Hyperparameter Optimization}
\label{subsec:hpo_tpe}

For each dataset $D \in \{D^{\mathrm{ov}}, D^{\mathrm{co}}\}$ we perform centralized \ac{HPO} using the Tree structured Parzen Estimator (\ac{TPE}) algorithm~\citep{Bergstra2011}. Let $\lambda$ denote a hyperparameter configuration (Equation~\ref{eq:tpe}),
\begin{equation}\label{eq:tpe}
    \lambda = (\eta, o, b),
\end{equation}
where $\eta \in \Lambda_{\eta}$ is the learning rate, $o \in \Lambda_{o}$ is the optimizer (e.g., SGD, Adam), and $b \in \Lambda_{b}$ is the batch size. Given a configuration $\lambda$, we train the model on the training split of $D$ and evaluate the resulting parameters on the validation split, obtaining a scalar performance metric $f(\lambda; D)$. The goal is to solve the Equation~\ref{eq:tpe-objective}:
\begin{equation}\label{eq:tpe-objective}
    \lambda^{\star}(D) = \arg\max_{\lambda \in \Lambda} f(\lambda; D),
\end{equation}
where $\Lambda = \Lambda_{\eta} \times \Lambda_{o} \times \Lambda_{b}$ is the search space.

The \ac{TPE} algorithm models the conditional density $p(\lambda \mid f)$ as a mixture of two densities over ``good'' and ``bad'' configurations and selects new candidates that maximize the expected improvement~\citep{Bergstra2011, Rodrigues2020}. After a fixed budget of evaluations, we obtain two dataset specific optima defined in Equation~\ref{eq:tpe-optima}:
\begin{equation}\label{eq:tpe-optima}
    \lambda^{\star}_{\mathrm{ov}} = \lambda^{\star}(D^{\mathrm{ov}}), 
    \qquad
    \lambda^{\star}_{\mathrm{co}} = \lambda^{\star}(D^{\mathrm{co}}).
\end{equation}

\begin{table}[!ht]
\centering
\caption{Hyperparameter Search Space.}
\label{tab:hyperparameters}
\renewcommand{\arraystretch}{1.2}
\resizebox{\columnwidth}{!}{
\begin{tabular}{ccc}
\hline
\textbf{Hyperparameter} & \textbf{Distribution} & \textbf{Search Space} \\ \hline
\ac{LR} &  Log-Uniform & $1 \times 10^{-5}$, $1 \times 10^{-3}$ \\
Batch Size & Categorical & $\{16, 32, 64\}$ \\
Optimizer & Categorical & $\{\text{Adam, SGD}\}$ \\ \hline
\end{tabular}
}
\end{table}

These configurations are later transferred to the federated phase and used as baselines. The search space for \ac{HPO} was defined and explored by \ac{TPE}, as detailed in Table \ref{tab:hyperparameters}. The hyperparameters tuned in this study were as follows:

\begin{itemize}
\item \textbf{Learning rate:} regulates the magnitude of each update applied to the model parameters during gradient-based optimization. We defined its search space on a logarithmic scale, which is a common choice for quantities that span several orders of magnitude, enabling the optimizer to probe the model’s sensitivity to different \ac{LR} values more effectively~\citep{Goodfellow2016, Rodrigues2020}.

\item \textbf{Batch size:} specifies how many training samples are used to compute each gradient update. It was treated as a discrete variable selected from a set of candidate values as it affects both the convergence behavior and computational efficiency~\citep{Goodfellow2016, Rodrigues2020}.

\item \textbf{Optimizer:} the optimization algorithm that updates the network weights to reduce the loss function. We modeled the optimizer choice as a categorical hyperparameter and allowed TPE to select the most suitable option for each configuration~\citep{Goodfellow2016}.
\end{itemize}

\subsection{Combined Hyperparameter Aggregation Heuristic}
\label{subsec:combined_heuristic}

The main contribution of this study is a simple yet explicit aggregation heuristic that combines dataset-specific optima into a single combined hyperparameter configuration. We aim to turn the optima obtained for distinct cancer tasks into a transparent candidate for generalizable hyperparameters in \ac{non-IID} \ac{FL}.

Let the optimal configurations found by \ac{TPE} on the ovarian and colon datasets be given according to Equation~\ref{eq:lambda_opt_datasets}:
\begin{equation}
    \lambda^{\star}_{\mathrm{ov}} = (\eta_{\mathrm{ov}}, o_{\mathrm{ov}}, b_{\mathrm{ov}}),
    \qquad
    \lambda^{\star}_{\mathrm{co}} = (\eta_{\mathrm{co}}, o_{\mathrm{co}}, b_{\mathrm{co}}),
    \label{eq:lambda_opt_datasets}
\end{equation}
where $\eta$ is the learning rate, $o$ is the optimizer, and $b$ is the batch size. Based on the two optima, we define the combined configuration $\bar{\lambda}$ as shown in Equation~\ref{eq:lambda_bar_def}.
\begin{equation}
    \bar{\lambda} = (\bar{\eta}, \bar{o}, \bar{b}),
    \label{eq:lambda_bar_def}
\end{equation}
With the learning rate given by the arithmetic mean, as defined in Equation~\ref{eq:eta_bar}:
\begin{equation}
    \bar{\eta} = \frac{\eta_{\mathrm{ov}} + \eta_{\mathrm{co}}}{2},
    \label{eq:eta_bar}
\end{equation}
and the optimizer and batch size obtained as modal choices among the dataset specific optima:
\begin{equation}
    \bar{o} = \operatorname{mode}\{o_{\mathrm{ov}}, o_{\mathrm{co}}\},
    \qquad
    \bar{b} = \operatorname{mode}\{b_{\mathrm{ov}}, b_{\mathrm{co}}\}.
    \label{eq:mode_ob}
\end{equation}
In case of a tie in Equation~\eqref{eq:mode_ob} (for example, different optimizers selected for each dataset with equal frequency), we break ties by choosing the value associated with the higher validation F1 score in the centralized phase. 
Algorithm~\ref{alg:combined_hpo} summarizes the construction of $\bar{\lambda}$.

\begin{algorithm}[!ht]
\caption{Combined hyperparameter configuration from dataset specific optima}

\label{alg:combined_hpo}
\begin{algorithmic}[1]
\REQUIRE Optimal configurations $\lambda^{\star}_{\mathrm{ov}} = (\eta_{\mathrm{ov}}, o_{\mathrm{ov}}, b_{\mathrm{ov}})$ and $\lambda^{\star}_{\mathrm{co}} = (\eta_{\mathrm{co}}, o_{\mathrm{co}}, b_{\mathrm{co}})$, validation scores $f(\lambda^{\star}_{\mathrm{ov}}; D^{\mathrm{ov}})$ and $f(\lambda^{\star}_{\mathrm{co}}; D^{\mathrm{co}})$
\STATE Compute averaged learning rate according to Equation~\eqref{eq:eta_bar}:
\[
    \bar{\eta} \leftarrow \frac{\eta_{\mathrm{ov}} + \eta_{\mathrm{co}}}{2}.
\]
\STATE Set candidate sets for optimizer and batch size:
\[
    \mathcal{O} \leftarrow \{o_{\mathrm{ov}}, o_{\mathrm{co}}\}, \quad
    \mathcal{B} \leftarrow \{b_{\mathrm{ov}}, b_{\mathrm{co}}\}.
\]
\STATE Compute $\bar{o} \leftarrow \operatorname{mode}(\mathcal{O})$; if there is a tie, choose the optimizer associated with the higher validation F1 score.
\STATE Compute $\bar{b} \leftarrow \operatorname{mode}(\mathcal{B})$; if there is a tie, choose the batch size associated with the higher validation F1 score.
\STATE \textbf{return} $\bar{\lambda} = (\bar{\eta}, \bar{o}, \bar{b})$ as in Equation~\eqref{eq:lambda_bar_def}.
\end{algorithmic}
\end{algorithm}

In the federated setting, we therefore evaluate three hyperparameter schemes: (i) $\lambda^{\star}_{\mathrm{ov}}$ (ovary optimized), (ii) $\lambda^{\star}_{\mathrm{co}}$ (colon optimized), and (iii) $\bar{\lambda}$ (combined), defined in Equations~\eqref{eq:lambda_opt_datasets} and~\eqref{eq:lambda_bar_def}. All other components of the pipeline were fixed. This design isolates the impact of hyperparameter selection on federated performance and frames the combined configuration as a strategy across diverse datasets in \ac{non-IID} federated cancer imaging scenarios.

\section{\uppercase{Results and Discussion}}\label{sec:results}

We carried out all experiments on a Dell Precision 5860 workstation equipped with an Intel Xeon w3-2435 processor (8 CPU cores at 3.1~GHz), 64~GB of DDR5 memory, and an NVIDIA RTX~4000 Ada Generation GPU with 20~GB of VRAM. We implemented the training, validation, and testing using the Flower~\citep{Beutel2020} framework, and adopted FedAvg~\citep{McMahan2017} as the aggregation algorithm.

\ac{HPO} for each \ac{CNN} is performed centrally with \ac{TPE}, using the Hyperopt~\citep{Bergstra2015} library. We construct the combined configuration $\bar{\lambda}$ via the proposed aggregation heuristic. In the federated experiments, we keep the model architectures, data partitions, and number of local epochs and communication rounds fixed, and vary only the hyperparameter scheme (colon-optimized, ovary-optimized, or combined). 

\subsection{Centralized HPO}
\label{subsec:results_hpo}

We first investigate the behavior of each \ac{CNN} under centralized training, using \ac{TPE} to obtain dataset specific hyperparameter optima. This experiment provides two baselines: it characterizes how well each architecture fits the colon and ovarian tasks in isolation, and it yields the colon optimized and ovary optimized configurations that will later be transferred to the federated setting and combined through the heuristic. Table~\ref{tab:hpo_centralized} summarizes the best batch size, optimizer, and validation loss found by TPE for each \ac{CNN} and dataset, while the corresponding learning rates are reported in Table~\ref{tab:lr_schemes}. 

\begin{table}[!ht]
\centering
\caption{Best batch size, optimizer, and validation loss found by TPE.}
\renewcommand{\arraystretch}{1.25}
\resizebox{\columnwidth}{!}{
\label{tab:hpo_centralized}
\begin{tabular}{@{}llccc@{}}
\toprule
\textbf{Dataset}       & \textbf{Model}  & \textbf{Batch size} & \textbf{Optimizer} & \textbf{Best val. loss} \\ \midrule
\multirow{6}{*}{Colon} & AlexNet         & 16                  & Adam               & 3.71 $\times 10^{-4}$   \\
                       & ResNet-50       & 32                  & Adam               & 1.39 $\times 10^{-4}$   \\
                       & ResNet-34       & 64                  & Adam               & 1.86 $\times 10^{-4}$   \\
                       & ResNet-18       & 32                  & Adam               & 2.49 $\times 10^{-5}$   \\
                       & SqueezeNet 1.1  & 16                  & Adam               & 1.57 $\times 10^{-4}$   \\
                       & EfficientNet-B0 & 32                  & Adam               & 4.59 $\times 10^{-4}$   \\ \midrule
\multirow{6}{*}{Ovary} & AlexNet         & 64                  & Adam               & 2.00 $\times 10^{-1}$   \\
                       & ResNet-50       & 64                  & Adam               & 1.84 $\times 10^{-1}$   \\
                       & ResNet-34       & 64                  & Adam               & 1.83 $\times 10^{-1}$   \\
                       & ResNet-18       & 32                  & Adam               & 2.16 $\times 10^{-1}$   \\
                       & SqueezeNet 1.1  & 16                  & SGD                & 2.96 $\times 10^{-1}$   \\
                       & EfficientNet-B0 & 64                  & Adam               & 1.02 $\times 10^{-1}$   \\ \bottomrule
\end{tabular}%
}
\end{table}

\begin{table}[!ht]
\centering
\caption{Best Learning rates found by TPE.}
\label{tab:lr_schemes}
\renewcommand{\arraystretch}{1.2}
\resizebox{\columnwidth}{!}{
\begin{tabular}{lccc}
\toprule
Model & $\eta_{\mathrm{co}}$ & $\eta_{\mathrm{ov}}$ & $\bar{\eta}$ \\
\midrule
AlexNet         
  & 1.28 $\times 10^{-4}$ & 9.06 $\times 10^{-5}$ & 1.09 $\times 10^{-4}$ \\
ResNet-50       
  & 2.22 $\times 10^{-4}$ & 2.49 $\times 10^{-5}$ & 1.23 $\times 10^{-4}$ \\
ResNet-34       
  & 1.59 $\times 10^{-4}$ & 3.86 $\times 10^{-5}$ & 9.88 $\times 10^{-5}$ \\
ResNet-18       
  & 2.90 $\times 10^{-4}$ & 3.12 $\times 10^{-5}$ & 1.61 $\times 10^{-4}$ \\
SqueezeNet 1.1  
  & 8.38 $\times 10^{-5}$ & 3.78 $\times 10^{-4}$ & 2.31 $\times 10^{-4}$ \\
EfficientNet-B0 
  & 5.99 $\times 10^{-4}$ & 9.64 $\times 10^{-4}$ & 7.82 $\times 10^{-4}$ \\
\bottomrule
\end{tabular}
}
\end{table}

On the colon task, all models converged with Adam and batch sizes between 16 and 64, with small learning rates across architectures. The lowest validation loss was obtained by ResNet18 (best loss $2.49\times 10^{-5}$ with batch size 32), followed by ResNet50 and SqueezeNet, which indicates that shallow residual networks and lightweight architectures can fit the colon dataset under centralized training.

On the ovarian task, \ac{TPE} again selected small learning rates and batch sizes between 16 and 64, but the ranking of architectures changed. EfficientNet-B0 achieved the lowest validation loss ($1.02\times 10^{-1}$ with batch size 64), while ResNet50 and ResNet34 exhibited similar losses around $1.8\times 10^{-1}$. This shift suggests that architectures with more expressive feature extractors, such as EfficientNet-B0, are better suited to capture the higher intra-class variability of ovarian histopathology compared to the colon dataset. These dataset-specific optima, $\lambda^{\star}_{\mathrm{ov}}$ and $\lambda^{\star}_{\mathrm{co}}$, provide the starting point for the federated experiments and for the combined configuration.

\begin{table*}[!ht]
\centering
\caption{Federated performance on the \ac{non-IID} scenario under three hyperparameter schemes.}
\label{tab:federated_results}
\renewcommand{\arraystretch}{1.3}
\resizebox{\linewidth}{!}{%
\begin{tabular}{lcccc|cccc|cccc}
\cline{2-13}
                         & \multicolumn{4}{c|}{\textbf{Colon-optimized}}             & \multicolumn{4}{c|}{\textbf{Ovary-optimized}}             & \multicolumn{4}{c}{\textbf{Combined}}                     \\ \hline
\textbf{Model}           & \textbf{Acc} & \textbf{Prec} & \textbf{Rec} & \textbf{F1} & \textbf{Acc} & \textbf{Prec} & \textbf{Rec} & \textbf{F1} & \textbf{Acc} & \textbf{Prec} & \textbf{Rec} & \textbf{F1} \\ \hline
\textbf{AlexNet}         & 0.905        & 0.906         & 0.905        & 0.896       & 0.890        & 0.885         & 0.890        & 0.882       & {\ul 0.905}  & {\ul 0.906}   & {\ul 0.905}  & {\ul 0.896} \\ 
\textbf{ResNet-18}       & 0.890        & 0.897         & 0.890        & 0.875       & 0.870        & 0.868         & 0.870        & 0.852       & {\ul 0.920}  & {\ul 0.921}   & {\ul 0.920}  & {\ul 0.914} \\
\textbf{ResNet-34}       & 0.910        & 0.919         & 0.910        & 0.899       & 0.920        & 0.921         & 0.920        & 0.914       & {\ul 0.920}  & {\ul 0.923}   & {\ul 0.920}  & {\ul 0.913} \\
\textbf{ResNet-50}       & 0.950        & 0.949         & 0.950        & 0.949       & {\ul 0.955}  & {\ul 0.957}   & {\ul 0.955}  & {\ul 0.953} & 0.935        & 0.940         & 0.935        & 0.930       \\
\textbf{SqueezeNet}      & 0.915        & 0.916         & 0.915        & 0.908       & {\ul 0.955}  & {\ul 0.954}   & {\ul 0.955}  & {\ul 0.954} & 0.915        & 0.914         & 0.915        & 0.910       \\
\textbf{EfficientNet-B0} & 0.890        & 0.903         & 0.890        & 0.873       & 0.865        & 0.884         & 0.865        & 0.836       & {\ul 0.900}  & {\ul 0.906}   & {\ul 0.900}  & {\ul 0.888} \\ \bottomrule
\end{tabular}%
}
\end{table*}
\subsection{Federated Performance Under Different Hyperparameter Schemes}
\label{subsec:results_federated}

In the federated setting we analyze how the three hyperparameter schemes affect performance on the joint \ac{non-IID} ovarian+colon scenario. Table~\ref{tab:federated_results} reports the federated results on the combined \ac{non-IID} ovarian+colon scenario for each \ac{CNN} and for the three hyperparameter schemes (see Section~\ref{subsec:combined_heuristic}). 

All experiments use 50 local epochs and 3 communication rounds, so differences in performance can be attributed to the choice of learning rate, optimizer, and batch size. For each \ac{CNN}, the best result across the three schemes is underlined in the table to highlight the most effective configuration.

For ResNet50, the ovary-optimized configuration yields the best federated performance, with accuracy $0.955$ and F1-score $0.953$, slightly outperforming the colon-optimized (F1 $0.949$) and combined (F1 $0.930$) settings. ResNet34 follows a similar pattern: F1-score of $0.914$ for the ovary-optimized configuration, compared to $0.899$ (colon-optimized) and $0.913$ (combined). 

\begin{figure}[!ht]
    \centering
    \includegraphics[width=\columnwidth]{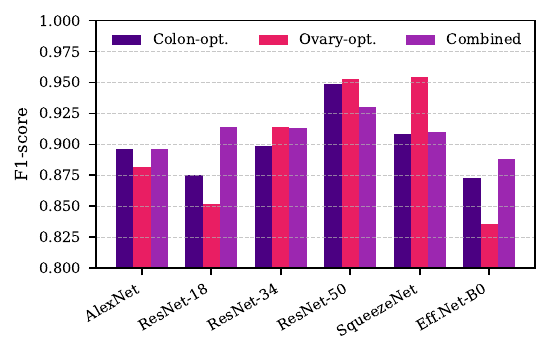}
    \caption{Federated F1-score for each \ac{non-IID} scenario evaluated.}
    \label{fig:federated_f1_schemes}
\end{figure}

In contrast, ResNet18 and EfficientNet-B0 benefit more from the combined configuration. ResNet18 achieves the highest F1-score ($0.914$) and accuracy ($0.920$), clearly improving over the colon-optimized ($\text{F1}=0.875$) and ovary-optimized ($\text{F1}=0.852$) settings. For EfficientNet-B0, the combined configuration attains an F1-score of $0.888$, outperforming both colon-optimized ($0.873$) and ovary-optimized ($0.836$) schemes. These trends are summarized in Figure~\ref{fig:federated_f1_schemes}.

AlexNet and SqueezeNet illustrate that no single dataset-specific optimum is universally dominant. AlexNet reaches its best F1-score under the colon-optimized configuration ($\text{F1}=0.896$), with the combined scheme matching this value and the ovary-optimized configuration slightly worse. For SqueezeNet, the ovary-optimized configuration with SGD achieves the highest F1-score ($0.954$), while the colon-optimized and combined schemes remain competitive around $0.908$–$0.910$. When averaging across all \ac{CNN}s, the combined configuration achieves the highest mean F1-score ($0.909$), compared to $0.900$ for the colon-optimized and $0.899$ for the ovary-optimized schemes. 


\section{\uppercase{Conclusion}}\label{sec:conclusion}

This study investigated whether hyperparameters optimized on one cancer histopathology dataset can generalize across \ac{non-IID} federated scenarios and tumor types. 
Centralized experiments showed that CNNs respond differently to the colon and ovarian tasks, with lightweight and moderately deep models (such as ResNet18) fitting the colon dataset particularly well, whereas EfficientNet-B0 achieved the best validation loss on ovarian images. In the federated \ac{non-IID} setting, the ovary-optimized configuration tended to favor deeper residual networks (e.g., ResNet50), whereas the combined configuration proved beneficial for lighter models, such as ResNet18 and EfficientNet-B0. When averaged across all architectures, the combined scheme achieved the highest mean F1-score and the lowest mean validation loss among the three hyperparameter schemes, indicating that the proposed aggregation heuristic can serve as a choice in a challenging \ac{non-IID} scenario.  


\section*{\uppercase{Acknowledgments}}
The authors gratefully acknowledges the financial support of FAPEMIG (Grant \#APQ00923-24) and the Institutional Program for Scientific Initiation Scholarships (PIBIC) at the Federal University of Viçosa (UFV). Andr\'e R. Backes gratefully acknowledges the financial support of CNPq (Grant \#302790/2024-1). 

\bibliographystyle{apalike}
\balance

{\small
\bibliography{refs}}

\begin{thebibliography}{}

\bibitem[Almelibari, 2025]{Almelibari2025}
Almelibari, A.~A. (2025).
\newblock Enhancing diabetes disease prediction and privacy preservation via federated learning and pso-wco optimization.
\newblock {\em Statistics, Optimization \& Information Computing}, 14:2297--2311.

\bibitem[Barbosa et~al., 2025]{Barbosa2025}
Barbosa, G. V.~G., Ferreira~Rodrigues, L.~G., F.~Rodrigues~Moreira, L., and Backes, A.~R. (2025).
\newblock {Federated Learning in Breast Cancer Diagnosis}.
\newblock {\em Revista de Informática Teórica e Aplicada}, 32(1):173–179.

\bibitem[Bergstra et~al., 2011]{Bergstra2011}
Bergstra, J., Bardenet, R., Bengio, Y., and K\'{e}gl, B. (2011).
\newblock {Algorithms for Hyper-Parameter Optimization}.
\newblock In {\em Advances in Neural Information Processing Systems}, volume~24. Curran Associates, Inc.

\bibitem[Bergstra et~al., 2015]{Bergstra2015}
Bergstra, J., Komer, B., Eliasmith, C., Yamins, D., and Cox, D.~D. (2015).
\newblock {Hyperopt: a Python library for model selection and hyperparameter optimization}.
\newblock {\em Computational Science \& Discovery}, 8(1):014008.

\bibitem[Beutel et~al., 2020]{Beutel2020}
Beutel, D.~J., Topal, T., Mathur, A., Qiu, X., Fernandez-Marques, J., Gao, Y., Sani, L., Li, K.~H., Parcollet, T., De~Gusm{\~a}o, P. P.~B., et~al. (2020).
\newblock Flower: A friendly federated learning research framework.
\newblock {\em arXiv preprint arXiv:2007.14390}.

\bibitem[Bray et~al., 2024]{Bray2024}
Bray, F., Laversanne, M., Sung, H., Ferlay, J., Siegel, R.~L., Soerjomataram, I., and Jemal, A. (2024).
\newblock {Global cancer statistics 2022: GLOBOCAN estimates of incidence and mortality worldwide for 36 cancers in 185 countries}.
\newblock {\em CA: A Cancer Journal for Clinicians}, 74(3):229--263.

\bibitem[Goodfellow et~al., 2016]{Goodfellow2016}
Goodfellow, I., Bengio, Y., and Courville, A. (2016).
\newblock {\em {Deep Learning}}.
\newblock MIT Press.
\newblock \url{http://www.deeplearningbook.org}.

\bibitem[Guan et~al., 2024]{Guan2024}
Guan, H., Yap, P.-T., Bozoki, A., and Liu, M. (2024).
\newblock {Federated learning for medical image analysis: A survey}.
\newblock {\em Pattern Recognition}, 151:110424.

\bibitem[He et~al., 2016]{He2016}
He, K., Zhang, X., Ren, S., and Sun, J. (2016).
\newblock {Deep Residual Learning for Image Recognition}.
\newblock In {\em 2016 IEEE Conference on Computer Vision and Pattern Recognition (CVPR)}, pages 770--778.

\bibitem[Khan et~al., 2024]{Khan2024}
Khan, S., Nosheen, F., Naqvi, S. S.~A., Jamil, H., Faseeh, M., Ali~Khan, M., and Kim, D.-H. (2024).
\newblock Bilevel hyperparameter optimization and neural architecture search for enhanced breast cancer detection in smart hospitals interconnected with decentralized federated learning environment.
\newblock {\em IEEE Access}, 12:63618--63628.

\bibitem[Krizhevsky et~al., 2012]{Krizhevsky2012}
Krizhevsky, A., Sutskever, I., and Hinton, G.~E. (2012).
\newblock {ImageNet Classification with Deep Convolutional Neural Networks}.
\newblock In {\em Advances in Neural Information Processing Systems 25}, pages 1097--1105. Curran Associates, Inc.

\bibitem[Kundroo and Kim, 2024]{Kundroo2024}
Kundroo, M. and Kim, T. (2024).
\newblock Demystifying impact of key hyper-parameters in federated learning: A case study on cifar-10 and fashionmnist.
\newblock {\em IEEE Access}, 12:120570--120583.

\bibitem[Kuo et~al., 2023]{Kuo2023}
Kuo, K., Thaker, P., Khodak, M., Nguyen, J., Jiang, D., Talwalkar, A., and Smith, V. (2023).
\newblock {On Noisy Evaluation in Federated Hyperparameter Tuning}.
\newblock In {\em Proceedings of Machine Learning and Systems}, volume~5, pages 127--144. Curan.

\bibitem[Li et~al., 2025]{Li2025}
Li, H., Govindarajan, V., Ang, T.~F., Shaikh, Z.~A., Ksibi, A., Chen, Y.-L., Ku, C.~S., Leong, M.~C., Shabaruddin, F.~H., Ishak, W. Z.~W., and Por, L.~Y. (2025).
\newblock Mspo: A machine learning hyperparameter optimization method for enhanced breast cancer image classification.
\newblock {\em DIGITAL HEALTH}, 11:20552076251361603.

\bibitem[McMahan et~al., 2017]{McMahan2017}
McMahan, B., Moore, E., Ramage, D., Hampson, S., and Arcas, B. A.~y. (2017).
\newblock {Communication-Efficient Learning of Deep Networks from Decentralized Data}.
\newblock In Singh, A. and Zhu, J., editors, {\em Proceedings of the 20th International Conference on Artificial Intelligence and Statistics}, volume~54, pages 1273--1282. PMLR.

\bibitem[Morgan et~al., 2022]{Morgan2022}
Morgan, E., Arnold, M., Gini, A., Lorenzoni, V., Cabasag, C.~J., Laversanne, M., Vignat, J., Ferlay, J., Murphy, N., and Bray, F. (2022).
\newblock Global burden of colorectal cancer in 2020 and 2040: incidence and mortality estimates from {GLOBOCAN}.
\newblock {\em Gut}, 72(2):338--344.

\bibitem[Rodrigues et~al., 2020]{Rodrigues2020}
Rodrigues, L.~F., Naldi, M.~C., and Mari, J.~F. (2020).
\newblock {Comparing convolutional neural networks and preprocessing techniques for HEp-2 cell classification in immunofluorescence images}.
\newblock {\em Computers in Biology and Medicine}, 116:103542.

\bibitem[Rodrigues and Backes, 2025]{RodriguesMoreira2025jdi}
Rodrigues, Moreira, L.~F. and Backes, A.~R. (2025).
\newblock Ensemble of handcrafted and learned features for colorectal cancer classification.
\newblock {\em Journal of Imaging Informatics in Medicine}.

\bibitem[Rodrigues et~al., 2025]{Leonardo2025}
Rodrigues, L. G.~F., V.~Gomes~Barbosa, G., Moreira, R., Rodrigues~Moreira, L.~F., and Backes, A.~R. (2025).
\newblock Medical image classification with privacy: Centralized and federated learning comparison.
\newblock {\em Revista de Informática Teórica e Aplicada}, 32(1):180–187.

\bibitem[Rodrigues~Moreira et~al., 2024]{Moreira2025}
Rodrigues~Moreira, L.~F., Moreira, R., Martins, E.~T., Jansen, V.~F., Lima, Y.~S., Rodrigues, L. G.~F., Travençolo, B. A.~N., and Backes, A.~R. (2024).
\newblock {Maximizing the Power of Cognitive Services with an AI-as-a-Service Architecture for Seamless Delivery}.
\newblock In {\em 2024 IEEE 13th International Conference on Cloud Networking (CloudNet)}, pages 1--8, Rio de Janeiro, Brazil. IEEE.

\bibitem[Roshandel et~al., 2024]{Roshandel2024}
Roshandel, G., Ghasemi-Kebria, F., and Malekzadeh, R. (2024).
\newblock {Colorectal Cancer: Epidemiology, Risk Factors, and Prevention}.
\newblock {\em Cancers}, 16(8).

\bibitem[Subramanian et~al., 2022]{Subramanian2022}
Subramanian, M., Rajasekar, V., V.~E., S., Shanmugavadivel, K., and Nandhini, P.~S. (2022).
\newblock {Effectiveness of Decentralized Federated Learning Algorithms in Healthcare: A Case Study on Cancer Classification}.
\newblock {\em Electronics}, 11(24).

\bibitem[Tamanini et~al., 2025]{Tamanini2025}
Tamanini, B., Sousa, V., Rodrigues, L., Oliveira, D., Dias, C., Cruz, L., Diniz, J., and Júnior, L.~S. (2025).
\newblock {Classificação de Carcinoma Endometrioide de Ovário por Transformação de Esquema de Cor e Radiomics em Imagens Histopatológicas}.
\newblock In {\em Anais do XXV Simpósio Brasileiro de Computação Aplicada à Saúde}, pages 68--79, Porto Alegre, RS, Brasil.

\bibitem[Tan and Le, 2019]{Tan2019}
Tan, M. and Le, Q.~V. (2019).
\newblock Efficientnet: Rethinking model scaling for convolutional neural networks.
\newblock {\em CoRR}, abs/1905.11946.

\bibitem[Yang et~al., 2025]{Yang2025}
Yang, J., Qin, H., Wang, J., Yee, P.~L., Prajapat, S., Kumar, G., Balusamy, B., Bashir, A.~K., and Omar, M. (2025).
\newblock {IoT-Driven Skin Cancer Detection: Active Learning and Hyperparameter Optimization for Enhanced Accuracy}.
\newblock {\em IEEE Journal of Biomedical and Health Informatics}, pages 1--11.

\end{thebibliography}



\end{document}